\documentclass[10pt]{article}
\usepackage[letterpaper]{geometry}
\usepackage{hicss}
\usepackage{times}
\usepackage[none]{hyphenat}
\usepackage{url}
\usepackage{latexsym}
\usepackage{graphicx}
\graphicspath{{images/}}
\usepackage[
    style=apa,
  ]{biblatex}
\usepackage{amssymb}
\usepackage{amsmath}
\usepackage{times}
\usepackage{soul}
\usepackage{url}
\usepackage[hidelinks]{hyperref}
\usepackage[utf8]{inputenc}
\usepackage[small]{caption}
\usepackage{subcaption}
\usepackage{graphicx}
\usepackage{booktabs}
\usepackage{algorithm}
\usepackage{algorithmic}
\usepackage{color,soul}
\usepackage{cleveref}
\usepackage{multirow} 
\usepackage{array}
\newcolumntype{H}{>{\setbox0=\hbox\bgroup}c<{\egroup}@{}}

\title{Utilizing Data Fingerprints for Privacy-Preserving Algorithm Selection in Time Series Classification: Performance and Uncertainty Estimation on Unseen Datasets}

\author{Lars B{\"o}cking$^*$\\
University of Bayreuth\\
\& Fraunhofer FIT\\
 {\underline{lars.boecking@uni-bayreuth.de}} \\ \And
 Leopold M{\"u}ller$^*$\\
University of Bayreuth\\
\& Fraunhofer FIT\\
 {\underline{leopold.mueller@uni-bayreuth.de} } \\ \And
  Niklas K{\"u}hl\\
University of Bayreuth\\
\& Fraunhofer FIT\\
 {\underline{kuehl@uni-bayreuth.de} } \\ \\
}

\date{}

\begin{document}
\maketitle
\def\thefootnote{*}\footnotetext{These authors contributed equally to this work}\def\thefootnote{\arabic{footnote}}

\begin{abstract}
The selection of algorithms is a crucial step in designing AI services for real-world time series classification use cases. 
Traditional methods such as neural architecture search, automated machine learning, combined algorithm selection, and hyperparameter optimizations are effective but require considerable computational resources and necessitate access to all data points to run their optimizations.
In this work, we introduce a novel data fingerprint that describes any time series classification dataset in a privacy-preserving manner and provides insight into the algorithm selection problem without requiring training on the (unseen) dataset. 
By decomposing the multi-target regression problem, only our data fingerprints are used to estimate algorithm performance and uncertainty in a scalable and adaptable manner. 
Our approach is evaluated on the 112 University of California riverside benchmark datasets, demonstrating its effectiveness in predicting the performance of 35 state-of-the-art algorithms and providing valuable insights for effective algorithm selection in time series classification service systems, improving a naive baseline by 7.32\% on average in estimating the mean performance and 15.81\% in estimating the uncertainty. 
\end{abstract}

\subsubsection*{Keywords:}

time series classification, performance estimation, quantification of model risk

\section{Introduction}
\begin{figure}[tbh]
\includegraphics[width=\linewidth]{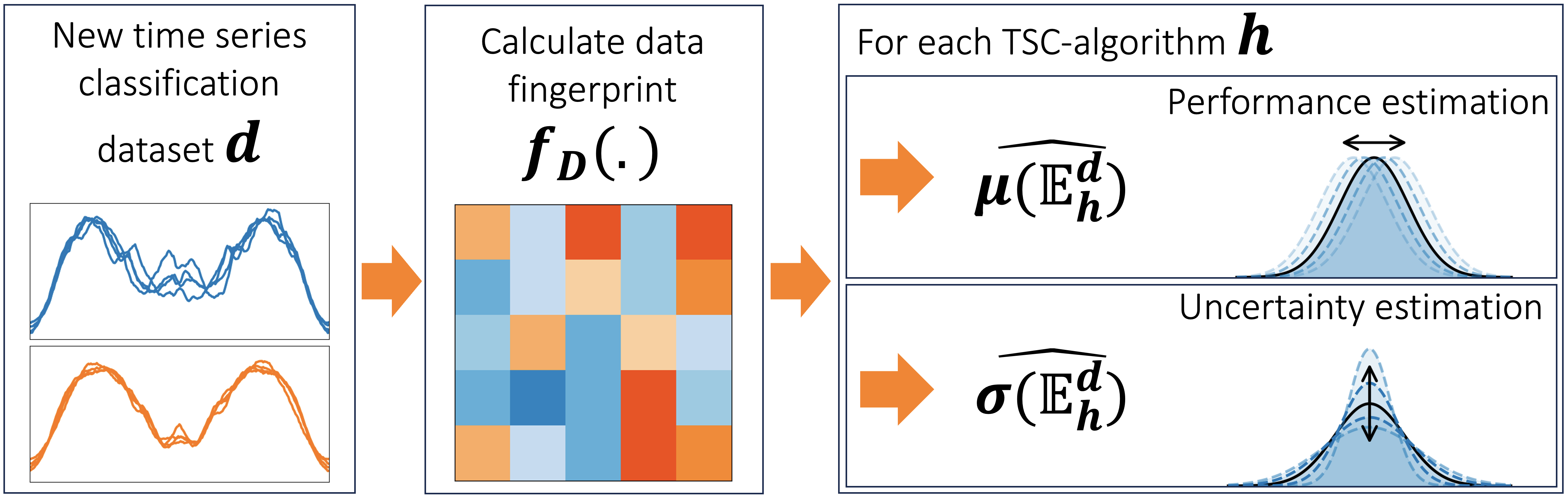}
  \caption{Approach for performance estimation in time series classification. Inspired by \cite{amini2020deep}.}
  \label{fig:concept_paper}
\end{figure}

Time series classification involves analyzing sequences of data points, indexed in time order, to categorize them into predefined classes. 
It is crucial in various services such as health record analysis \parencite{wang2022systematic}, predictive maintenance \parencite{ rudolph2020towards}, cyber-security \parencite{montazerishatoori2020detection}, and earthquake prediction \parencite{arul2021applications}, reflecting its wide-ranging impact in both scientific and practical applications.

\begin{figure}[tbh]
     \centering
     \begin{subfigure}[b]{0.49\textwidth}
         \centering
            \includegraphics[width=\linewidth]{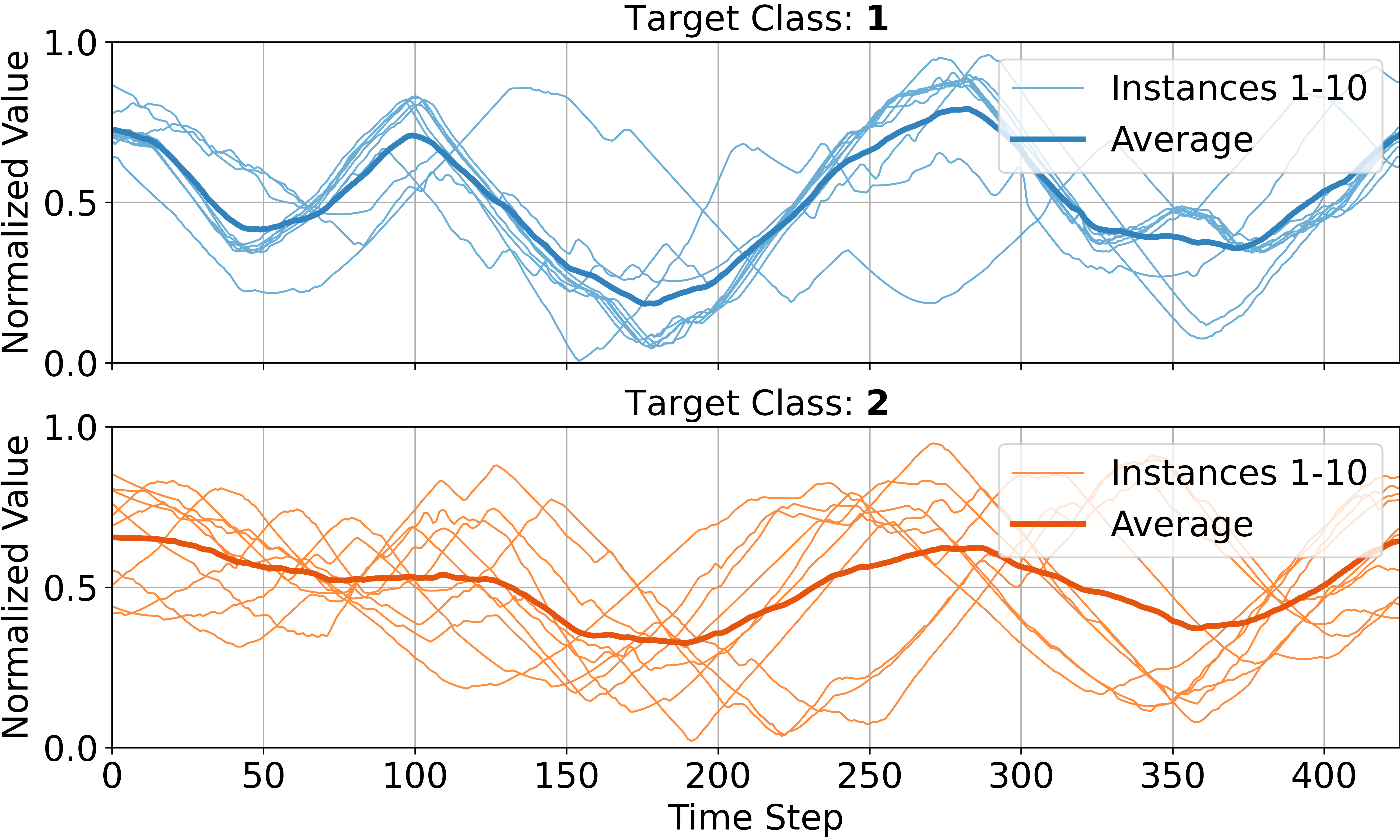}
            \caption{Sampled time series instances.}
            \label{fig:yoga_instances}
     \end{subfigure}
     \hfill
     \begin{subfigure}[b]{0.49\textwidth}
         \centering
            \includegraphics[width=\linewidth]{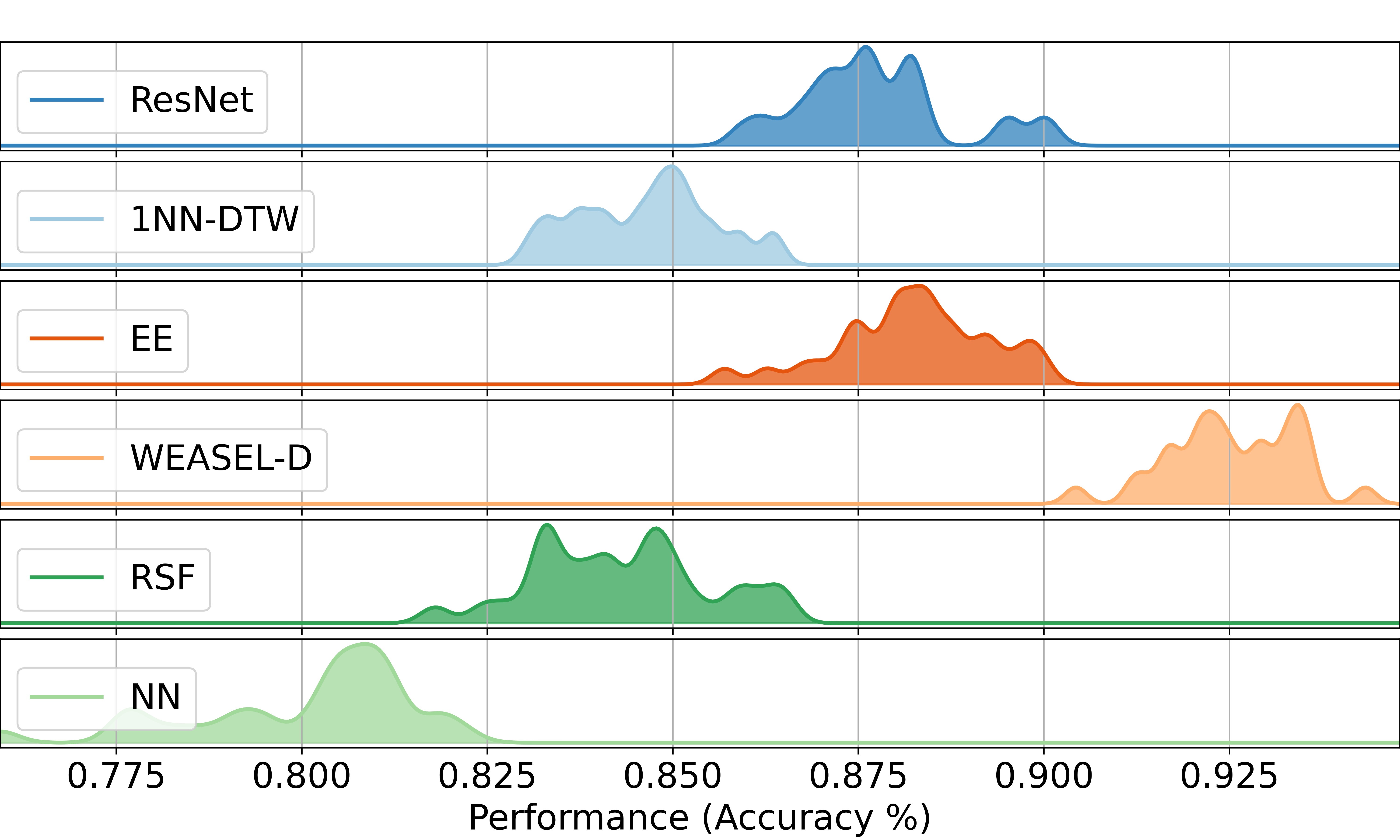}
            \caption{Histogram on the classification performance.}
            \label{fig:yoga_performance}
     \end{subfigure}
        \caption{Problem statement: Mapping the time series instances to the algorithm performance. \Cref{fig:yoga_instances}: Ten sampled time series instances of each target class and their averages in the \textit{Yoga}-dataset. \Cref{fig:yoga_performance}: Histogram of the classification performance of different algorithms on the \textit{Yoga}-dataset across 30 cross validation folds. Achieved accuracy on the x-axis and density on the y-axis.}
        \label{fig:problem_statement}
\end{figure}

The plethora of algorithms developed for time series classification presents a significant challenge: the selection of the most appropriate algorithm in a service to achieve the best performance for a specific dataset---and, therefore, the related domain problem---is a complex task \parencite{orevski2018data}.
This task is formalized by Rice and is known as the algorithm selection (AS) problem \parencite{algorithm_selection_problem}.
According to the ``no free lunch'' theorem, under certain assumptions, no algorithm uniformly dominates all others \parencite{no_free_lunch}. It is not known which algorithm will perform best on the new dataset.
Therefore, we present an approach to estimate these performances based on a data fingerprint, describing the dataset in an aggregated manner \Cref{fig:concept_paper}.
As a result, our approach allows us to differentiate variance in classification accuracies among different algorithms on real-world datasets, such as the UCR \textit{Yoga}-dataset \parencite{bagnall2017great}, an observation highlighted in \Cref{fig:yoga_performance}.

Organizations, researchers and service provider alike are constrained by limited resources—time, computation, data, and expertise.
Other methods, such as neural architecture search (NAS) \parencite{elsken2019nas}, automated machine learning (AutoML) \parencite{hutter2019automated, feurer2022auto}, transfer learning (TL) \parencite{lu2015transfer}, hyperparameter optimization (HPO) \parencite{hyperparam_2023}
and algorithm configuration \parencite{lindauer2015autofolio} engage in broad algorithmic experimentation or adaptation of pre-trained models for similar tasks.
These methods, despite their effectiveness, face substantial resource demands and complexity in selecting the most appropriate algorithm for possible deployment within an AI service \parencite{elsken2018efficient}. In addition, they necessitate access to data points to run their optimizations, compromising data privacy.
In scenarios where data privacy is a concern, the AI service provider needs a solution that allows for informed decision-making without requiring full access to the dataset. 
For a new, unseen time series classification dataset, the service provider wants to assess which state-of-the-art algorithm is most promising without training the algorithms themselves, running HPO or NAS. Instead, this can be achieved by analyzing the dataset's characteristics, enabling the service to act as an assistant in the AI development process while maintaining data privacy.

To address these shortcomings, we want to answer the following research questions (RQs):
\textbf{(RQ1)} How can diverse time series datasets be translated into a standardized input format to facilitate comparison and analysis?
\textbf{(RQ2)} To what extent can this standardized input format be used to estimate the expected performance of various state-of-the-art classification algorithms?
\textbf{(RQ3)} How can the uncertainty associated with the predicted algorithm performance on these standardized inputs be estimated?

In our work, we propose a data fingerprint to characterize datasets. We translate the algorithm selection (AS) problem into a multi-target regression problem that estimates algorithm performance and uncertainty, as illustrated in \Cref{fig:concept_paper}. We train various regressors on data fingerprints from benchmark datasets to predict the performance of time series classification algorithms on these benchmarks. Once trained, these regressors can be applied to new, unseen data fingerprints to predict how each algorithm within a service system will perform on the new datasets.
As a result, our approach effectively suggests the most suitable algorithm for any new dataset in a privacy-preserving manner, as only the data fingerprint is shared, not the actual data points. 
Therefore, a service provider does not need to access the dataset but can assist in the AI development process by suggesting the most promising classification algorithm---only by processing the fingerprint, which does not expose any data points. 
Our approach is highly customizable and can provide tailored suggestions by predicting other target variables in addition to accuracy and uncertainty to provide a basis for informed decision-making for AS in AI services.

Our contributions are threefold:

\begin{enumerate}
    \item We introduce novel data fingerprints to form feature maps for AS representing whole time series datasets. They capture the essential attributes of any time series dataset, making it easier to compare datasets and providing a standardized input for regression models. %

    \item We present a customizable approach that utilizes the fingerprints and benchmark results to decompose any multi-target regression problem related to AS. This includes estimating algorithm performance and its uncertainty in the work at hand---but can easily be adapted to other target objectives,to select the algorithm that will deliver the highest performance on a given dataset. Our approach enables efficient AS without the need for extensive training and testing of multiple algorithms, thus streamlining the process of achieving optimal model performance.

    \item We extensively experiment on 112 benchmark datasets and the expected performance of 35 state-of-the-art classification algorithms on unseen datasets. The results of our analysis provide insights for researchers and practitioners navigating AS in time series classification. Surpassing a naive baseline by an average of $7.32\%$ in estimating the mean performance and $15.81\%$ in estimating the uncertainty, it shows the approach's potential for future AS problems. We offer an out-of-the-box framework that predicts algorithm performance on any (unseen) dataset using the fingerprints.

\end{enumerate}

\section{Related Work}
In this section, we review various methods for algorithm selection, highlighting the computational challenges and privacy concerns associated with these methods to give an overview of the state-of-the-art in the field.

Algorithm selection: AS generally describes the selection of the most suitable algorithms for novel tasks \parencite{algorithm_selection_problem}. Unlike the broader approach that does not distinguish between general and machine learning-specific algorithms, our focus is squarely on the latter. We adopt a meta-learning perspective, utilizing machine learning algorithms not for direct selection but for estimating performance and uncertainty of potential algorithm choices. Based on these results, we use the predictions of the best algorithm selectors to make a final prediction about the expected performance and uncertainty. However, the predictions could also be used for the selection itself. For example, the algorithm with the highest performance could be selected. Other target values, such as the expected running time, could also be estimated and used as a basis for decision-making.
A distinction can be made between online and offline AS \parencite{degroote2017online}. Online AS describes the case where no training data is available in advance, and the selection is made iteratively. Our approach is to be considered offline AS.

Distinction from other methods: The selection of hyperparameters for an algorithm can also be optimized, which is referred to as HPO \parencite{hyperparam_2023}; if they are generalized for a set of tasks, this is called algorithm configuration \parencite{schede2022survey}. As we use benchmarks as a data basis, we assume that the benchmarks utilized already incorporate potential performance improvements achievable through HPO or algorithm configuration.
This is also an advantage over other methods such as NAS \parencite{elsken2019nas} or AutoML \parencite{hutter2019automated}, which have also been applied to the algorithm selection problem in time-series classification \parencite{10184850, 9643158}.
These methods are not only computationally intensive, as they attempt to find an optimal architecture or algorithm through targeted experimentation, but they also require full access to data points, compromising data privacy. There is ongoing research addressing the limitation of data privacy in NAS and AutoML \parencite{wang2022towards,zhang2021privacy,yan2022privacy}. These works, although primarily focused on domains other than time series classification, highlight the importance of privacy-preserving techniques in algorithm selection and model training. If both AS and HPO are carried out, it is named the combined algorithm selection and hyperparameter optimization (CASH) problem \parencite{thornton2013auto}. A large number of algorithms already exist for time series classification. Our work is intended to help estimate the performance and uncertainty for new datasets and thus support the selection process instead of searching for new architectures in a computationally intensive manner. TL \parencite{lu2015transfer} is also based on existing algorithms, which are adapted to the new task. However, here too, the expected performance is unknown in advance.

\section{Approach}
Consider an AI service that supports the development of time series classification solutions by recommending the most suitable algorithms based on data characteristics, enabling more informed AS and solutions tailored to the specific data of new clients.
While the current state-of-the-art time series classification algorithm can be employed, this approach often overlooks the complexities and nuances inherent in real-world data.
These algorithms typically rely on benchmark datasets that may not fully capture the intricacies of diverse datasets.
Instead, our approach identifies the most promising algorithm based on the unique characteristics of the new client's dataset, thereby delivering a truly smart service as defined by \cite{jussen2020smart}. 

To formalize our approach, we start by defining the time series classification problem, its performance assessment, and key concepts of our fingerprint aggregation.
For a given time series $x$, a time series classification algorithm predicts its target class $y$. 
Each task is represented by an individual dataset $d$ with an arbitrary number of instances $x^i$ to be classified.
Let $\mathbf{x}^{i,d}$ denote a $(1 \times T)$ vector of an univariate time series. 
The classification task is to learn $h(.)$: 
$x^{i,d}= \left( x_{1}^{i,d},\ldots x_{T}^{i,d} \right) \xrightarrow{h} y^{i,d}$
with: $d: \text{datasets, } 1 \ldots D, i: \text{instance, } 1 \ldots I, t: \text{time, } 1 \ldots T $. 

For an algorithm $h(.)$, its classification performance on a dataset $d$ is defined as $\mathbb{E}_{h}^{d} \left[ \mathbf{1}{\{y^{i,d} = h(x^{i,d})\}} \mid \tau \right]$, where $\tau$ relates to all training instances and their respective labels \parencite{hastie2009elements}. 
Predicting how the algorithms will perform on this task can provide insight into which algorithm should be chosen for the task.
To do this, we first translate the dataset into a standardized data fingerprint $f_D^d(.)$ of fixed size, so that it can be used as input to a regressor. Then we learn a mapping $f_D^d(.)\xrightarrow{}\mathbb{E}_{h}^{d}$ between the data fingerprint and the expected performance $\mathbb{E}_{h}^{d}$.

\begin{figure}[tbh]
\includegraphics[width=\linewidth]{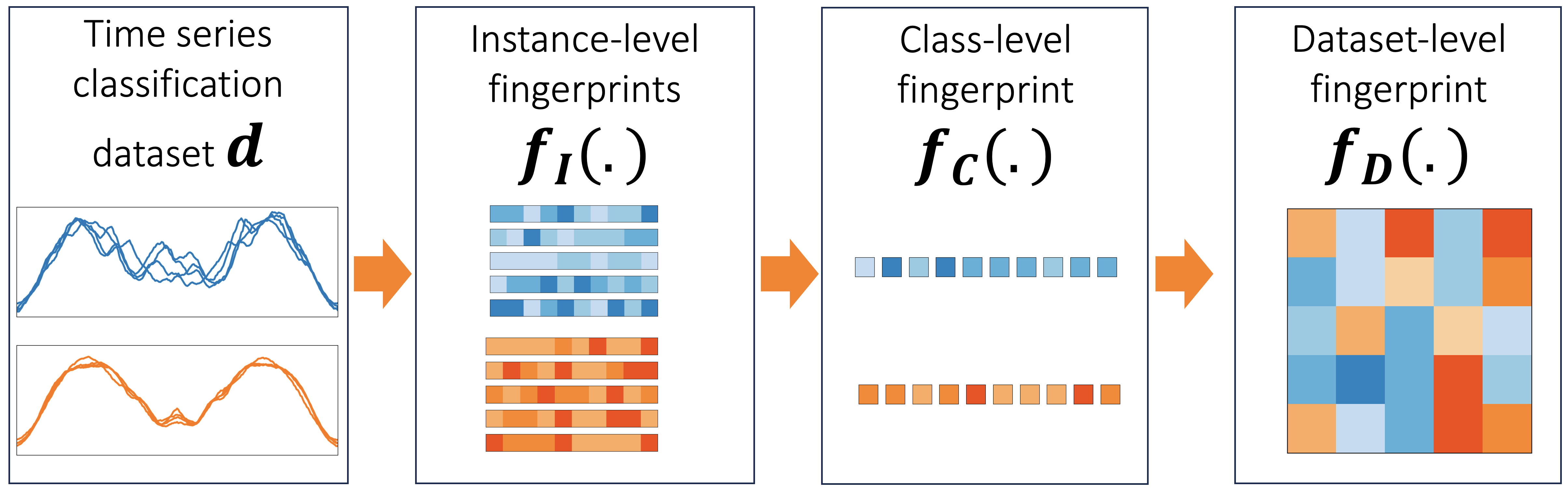}
  \caption{Approach step-wise aggregation function.}
  \label{fig:concept_aggregation_paper}
\end{figure}

More precisely, we map the data fingerprint to the mean $\mu{(\mathbb{E}_{h}^{d})}$ and the standard deviation $\sigma{(\mathbb{E}_{h}^{d})}$ of the performance, which occurs over multiple training runs, using k-fold cross-validation.
The standardized data fingerprint $f_D^d(.)$ is derived by aggregating information about the time series dataset on three levels, motivated by balancing the discriminating power of characteristics while consolidating information about individual time series into a vector of fixed size. 
This step-wise aggregation is shown in \Cref{fig:concept_aggregation_paper} and determines the structure of the remaining section: Instance-level fingerprint (see \Cref{sec:instance_characteristics}), 
class-level fingerprint (see \Cref{sec:class_characteristics}), and the dataset-level fingerprint (see \Cref{sec:dataset_characteristics}).

\subsection{Instance-level fingerprint} \label{sec:instance_characteristics}
On the first level, we want to describe each instance by representations of fixed size instead of the actual values themselves. Thus, we introduce an instance-level fingerprint.
This instance-level fingerprint is described by $f_I(.)$ and is calculated for each instance $x^{i,d}$ of a dataset $d$ separately:
$\forall i \in I: f_I(x^{i,d})$.
One exemplary representation could be to identify the deviation of change in one time step to the average deviation given by $\overline{\Delta x^{i,d}}$, such as $f_I(x^{i,d}) = \sqrt{\frac{1}{T-2} \sum\limits_{t=1}^{T-1} \left( x^{i,d}_{t+1} - x^{i,d}_t - \overline{\Delta x^{i,d}} \right)^2}$.

Other statistical measures can also be used to capture different aspects of the time series data.
The descriptive statistics are combined in a $(1 \times L_I)$ vector, where $L_I$ is the number of statistical measures used, to form the instance-level fingerprint $f_I(x^{i,d})$. An overview of the proposed statistical measures like Skewness $\gamma_{1}$ and Kurtosis $Kurt[X]$ can be found in the code of this work.

\subsection{Class-level fingerprint}\label{sec:class_characteristics}

On the next level, we want to aggregate the instance-level fingerprints of each class into class-level fingerprints of fixed size. So each class is assigned a class-level fingerprint, which results from transforming all instance-level fingerprints belonging to that class.

To aggregate the instance fingerprint for all instances of a given class $c$ in a dataset $d$, we first define the set of indexes related to a specific class as $I^{c} = \{i | y^{i,d} = c\}$. 
We can then derive $f_C(.)$ as:
$\forall c \in \mathcal{C}: f_C^c\left(\{f_I(x^{i,d})\}_{i \in I^{c}}\right)$.
One option is to calculate the average of a given fingerprint across all instances of a class, so $f^{c}_C(.) = \frac{1}{|I^{c}|} \sum_{i \in I^{c}} f_I\left(x^{i,d}\right)$.
Another option is to take the median value as a representation for the class instances.
Again, $f_C(.)$ can be independently selected from $f_I(.)$ or the later defined data fingerprint $f_D(.)$.
Our approach provides a generalizing concept that can be easily extended and adapted by choosing different aggregation functions.
\subsection{Dataset-level fingerprint}\label{sec:dataset_characteristics}

On the last level, we aggregate the previously calculated class-level fingerprint $f_C(.)$ and extend them by meta characteristics to form a standardized dataset-level fingerprint that describes any time series classification dataset as a function of $f_D(.)$, as:
$\forall d \in D: f_D^d\left( f_C^1\left( . \right),\ldots ,f_C^{|C|}\left(. \right) \right).$
One example aggregation function is the standard deviation of each class-level fingerprint across the available classes: $\sqrt{\frac{\sum_{c \in C} \left( f_{C}^c\left(.) \right) - \overline{f_{C}^c} \right)^2}{|C|}}$.

Besides the aggregated $f_C(.)$, we also add meta characteristics of our dataset, such as the total number of training and test instances, the length of each instance expressed as $||x^{i,d}||$, number of target classes. Moreover, the distribution of instances across classes is characterized by the minimum and maximum number of instances in any class, represented by $\min(||I_{c}||)$ and $\max(||I_{c}||)$, respectively as well as the average number of instances per class and the standard deviation of the number of instances per class.
This fixed size $(1 \times L_D)$ vector, where $L_D$ is determined by the dataset-level aggregation type and the number of meta characteristics, provides a comprehensive dataset description and serves as the input for our multi-target regression problem. Details on the decomposition of this multi-target regression are discussed in the following subsection.

\subsection{Performance estimation}\label{sec:performance_estimation}

There is no single algorithm that performs best on all available tasks---the ``no free lunch'' theorem \parencite{no_free_lunch}. Our approach addresses this by mapping our proposed fingerprint $f_D^d(.)$ to any multi-objective performance measures defined by the multi-target regression problem. 
It does so by decomposing the performance and its uncertainty, as well as the algorithm $h(.)$, learning a regressor $r(.)$ separately as shown in \Cref{fig:concept_paper} on page \pageref{fig:concept_paper}.

Motivated by the central limit theorem \parencite{rosenblatt1956central} and the asymptotic characteristics of k-folds \parencite{li2023asymptotics}, we derive the estimation of the performance and its uncertainty for our approach.
We learn a regressor $r(.)$ such that 
$f_D^d(.)\xrightarrow{r(.)}\widehat{\mu(\mathbb{E}_{h}^{d})}$, estimating the mean classification performance of algorithm $h(.)$ on a dataset $d$ as well as the observed standard deviation in performance $f_D^d(.)\xrightarrow{r(.)}\widehat{\sigma(\mathbb{E}_{h}^{d})}$. 
Note that our approach and code allow us to estimate various characteristics of an algorithm's performance, e.g., lower percentiles of the k-folds for estimating lower bounds in a risk-averse setting like earthquake prediction \parencite{arul2021applications}.
Details on the regressors $r(.)$ applied to the multi-target regression AS problem can be found in \Cref{sec:estimation_regressors}.

\section{Experiment \& results}

Our approach estimates the performance of an algorithm $h(.)$ on a dataset $d$, described by $\mathbb{E}_{h}^{d}$, solely through the computation of select characteristics that describe the dataset. To train and test this mapping, we need various datasets and the related performance of multiple classification algorithms.
We evaluate our approach on the $112$ univariate time series datasets established in the UCR classification benchmark~\parencite{bagnall2017great, dau2019ucr}. 
The performances are established by \cite{middlehurst2023bake} in their bake-off paper and available as part of the time series machine learning package \parencite{middlehurst2023bake}.
We run our evaluation on all $35$ algorithms $h(.)$ referenced in this most recent benchmark, such as BOSS~\parencite{schafer2015boss}, HC2~\parencite{middlehurst2021hive}, InceptionT~\parencite{ismail2020inceptiontime}, ROCKET~\parencite{dempster2020rocket}, among others~\footnote{1NN-DTW, BOSS~\parencite{schafer2015boss}, Catch22~\parencite{lubba2019catch22}, cBOSS~\parencite{middlehurst2019scalable}, CIF~\parencite{middlehurst2020canonical}, CNN~\parencite{ismail2019deep}, EE~\parencite{lines2015time}, FreshPRINCE~\parencite{middlehurst2022freshprince}, HC1~\parencite{bagnall2020usage}, Arsenal, DrCIF and HC2~\parencite{middlehurst2021hive}, Hydra and Hydra-MR~\parencite{dempster2023hydra}, InceptionT~\parencite{ismail2020inceptiontime}, Mini-R~\parencite{dempster2021minirocket}, MrSQM~\parencite{nguyen2022fast}, Multi-R~\parencite{tan2022multirocket}, PF~\parencite{lucas2019proximity}, RDST~\parencite{guillaume2022random}, ResNet~\parencite{wang2017time}, RISE~\parencite{flynn2019contract}, RIST, ROCKET~\parencite{dempster2020rocket}, RSF~\parencite{karlsson2016generalized}, RSTSF~\parencite{cabello2021fast}, ShapeDTW~\parencite{zhao2018shapedtw}, Signatures, STC~\parencite{hills2014classification}, STSF~\parencite{cabello2020fast}, The Temporal Dictionary Ensemble {(TDE)}~\parencite{middlehurst20temporal}, TS-CHIEF~\parencite{shifaz20ts}, TSF~\parencite{deng2013time}, TSFresh~\parencite{christ2018time}, WEASEL~\parencite{schafer17fast}, WEASEL-D.}.
For each dataset $d \in D$ we calculate the instance fingerprint $f_I(.)$, the class fingerprint $f_C(.)$, and finally accumulate the data fingerprint $f_D^d(.)$.
Our approach estimates the classification performance $\widehat{\mu(\mathbb{E}_{h}^{d})}$ and uncertainty $\widehat{\sigma(\mathbb{E}_{h}^{d})}$ of algorithms $h(.)$ based on this final fingerprint. 

We split the 112 datasets of the UCR benchmark $d \in [1,...,D]$ by a $.2/.2/.6$ train-validation-test split. For each of the individual datasets in $D_{train}$, $D_{validation}$ and $D_{test}$, we calculate their fingerprint $f_D^d(.)$ and pair them with the achieved performance of each classification algorithm $h$.

The performance regressors $r(.)$ are trained on the fingerprint and classification performances of $h(.)$ on all datasets in $D_{train}$. The regressors $r(.)$ are selected based on their accuracy in performance estimation of classification algorithms $h(.)$ on $D_{validation}$. We evaluate our approach by running the regressor $r(.)$ on the fingerprints of $D_{test}$ and compare the estimated performances and uncertainty to the benchmark results. 
The code of this work is publicly available\footnote{\url{https://github.com/LarsBoecking/time_series_fingerprint}}.
\subsection{Naive baseline}

We derive a naive baseline $\ddot{\mu}_{h}^d$ for the mean performance $\mathbb{E}_{h}^{d}$ of an algorithm $h$ on a dataset $d$ building upon the common concept of a single best solver \parencite{bischl2016aslib}.
We define $\ddot{\mu}_{h}^d = \frac{1}{|D_{train}|} \sum_{d \in D_{train}} \mu(\mathbb{E}_{h}^{d})$. It reflects the average performance of algorithm $h$ on $D_{train}$, the datasets used for training our performance estimator. 
Correspondingly a naive baseline $\ddot{\sigma}_{h}^d$ for the expected uncertainty in performance can be calculated by $\ddot{\sigma}_{h}^d = \frac{1}{|D_{train}|} \sum_{d \in D_{train}} \sigma(\mathbb{E}_{h}^{d})$. 
Our approach estimates $\widehat{\mu(\mathbb{E}_{h}^{d})}$ and $\widehat{\sigma(\mathbb{E}_{h}^{d})}$ for any $d \in D_{test}$ is benchmarked against this baseline.

\subsection{Fingerprints}

\begin{figure}[tbh]
    \centering
    \includegraphics[width=\linewidth]{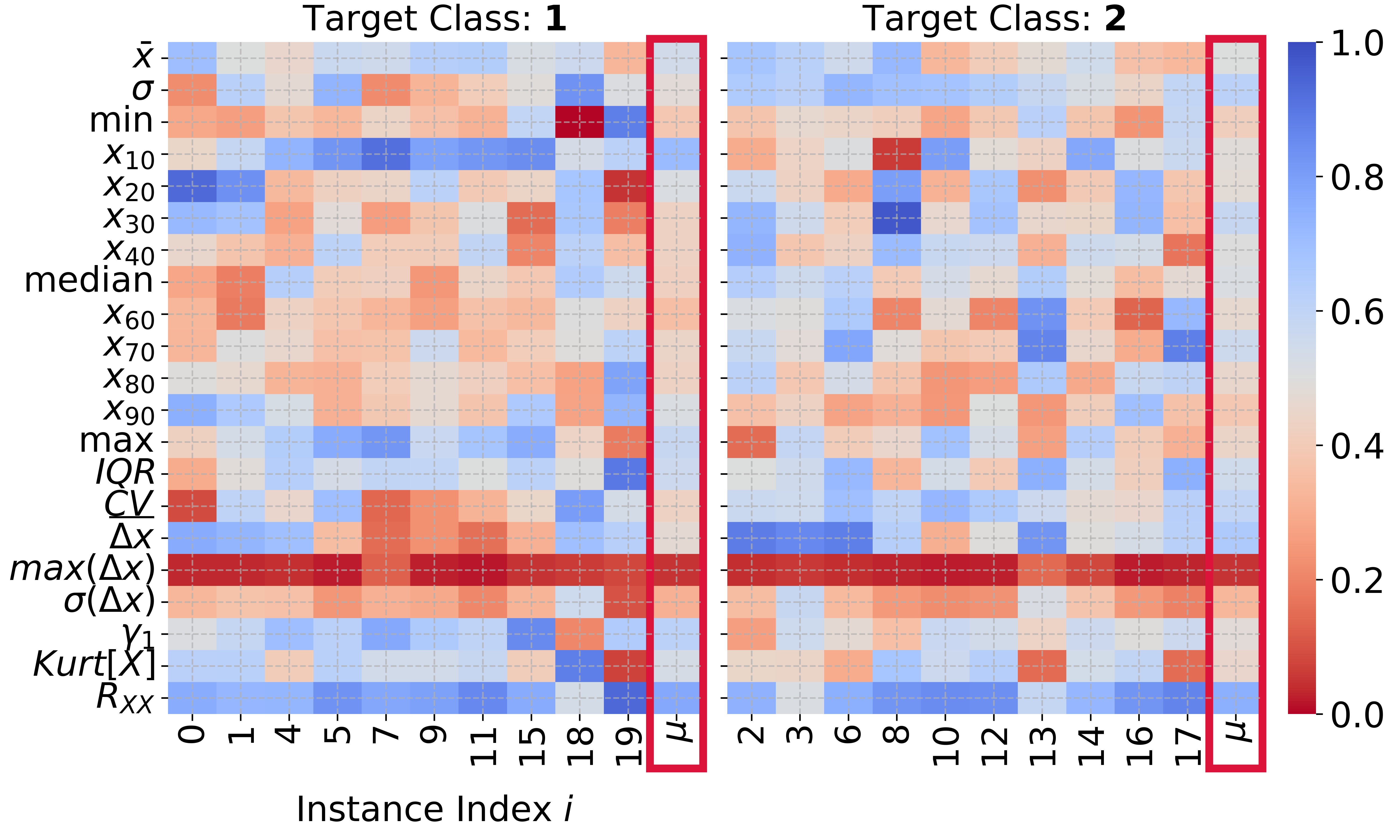}
    \caption {Instance fingerprint $f_I(x^{i,d})$ for ten sampled instances of each class in the \textit{Yoga}-dataset and the class aggregation $f_C(.)$ via $\mu$ aggregation. As indicated by their different values, Skewness $\gamma_{1}$ and Kurtosis $Kurt[X]$ are promising characteristics to differentiate individual instances as well as aggregated class fingerprints.}
    \label{fig:yoga_class_fingerprint}
\end{figure}

\textbf{Instance-level fingerprints}: The instance-level fingerprint $f_I(.)$ describes instances of any length by a fixed-size vector. 
Accurately differentiating each individual instance just by its fingerprint seems challenging, as shown in {\Cref{fig:yoga_instances}}. 
Still underlying patterns can be identified, e.g. instances of target class 1 in the \textit{Yoga}-dataset have higher Skewness $\gamma_{1}$ while instances of target class 2 have higher mean change $\overline{\Delta x^{i,d}}$, as shown in \Cref{fig:yoga_class_fingerprint}.

\textbf{Class-level fingerprint}: Our approach aggregates instance-level fingerprint $f_I(.)$ across all instances $I^c$ for each target class $c$.
\Cref{fig:yoga_class_fingerprint} provides a fingerprint for the first ten individual instances of each target class in the \textit{Yoga}-dataset, as well as their class-level aggregation $f_C(.)$.
Note that this visualization highlights the concept for a reduced number of instances (ten in this case). 
The actual class-level fingerprint $f_C(.)$ is aggregated on all instances $I^c$ in the training subset of the given dataset $d$. 
Still, only these ten instances result in a class fingerprint with certain distinguishing characteristics, e.g., Kurtosis and Skewness.

\textbf{Dataset-level fingerprint}: Finally, to build a fixed-size fingerprint that can be utilized to describe any time series classification task, the class-level fingerprints $f_C(.)$ are aggregated on dataset granularity. 
For the dataset-level aggregation, standard deviation, interquartile range, and the range between the minimum and maximum value are calculated.
Each of those fixed-sized fingerprints representing an individual dataset is then mapped to an algorithm performance $f_D^d(.)\xrightarrow{}\mathbb{E}_{h}^{d}$.

\subsection{Performance estimation for a given algorithm}
\begin{figure}[tbh]
    \centering
        \includegraphics[width=\linewidth]{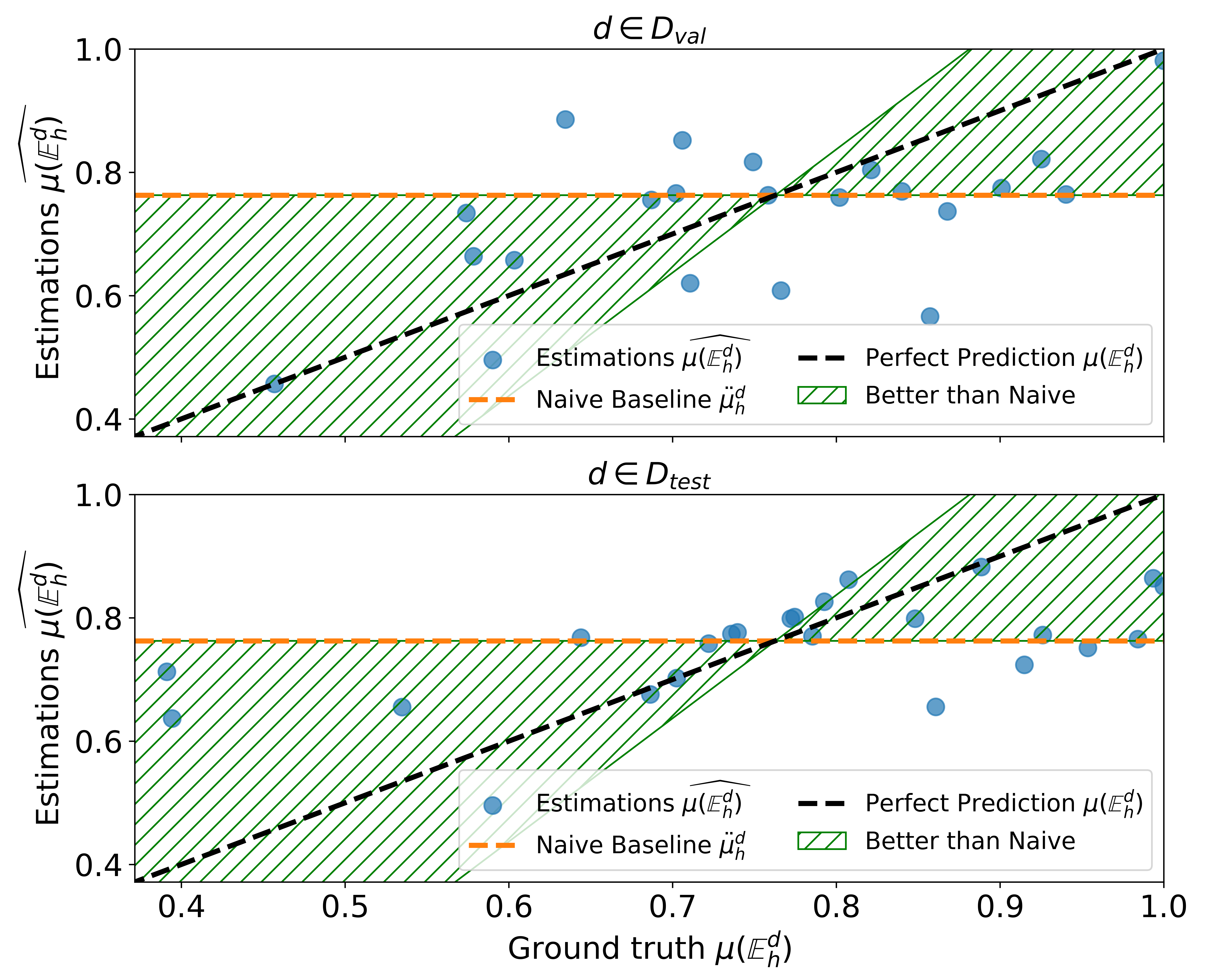}
        \caption{Ridge regression estimations $\widehat{\mu(\mathbb{E}_{h}^{d})}$ for $h$ 1NN-DTW, achieving an average improvement in MAE of $18.13\%$ on $D_{val}$ and $18.61\%$ on $D_{test}$ compared to $\ddot{\mu}_{h}^d$.}
        \label{fig:ridge_mean}
\end{figure}

\begin{figure}[tbh]
    \centering
        \centering
        \includegraphics[width=\linewidth]{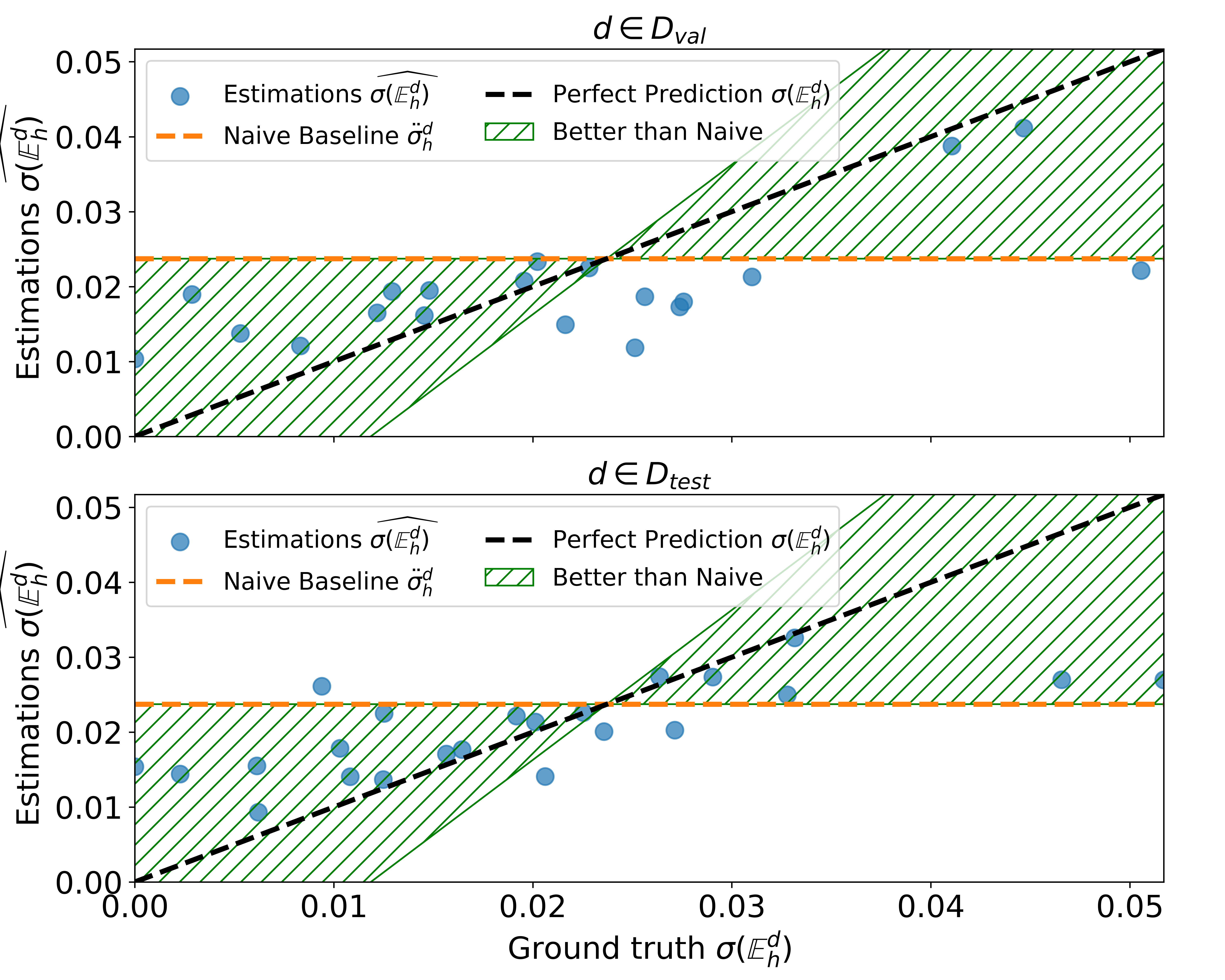}
        \caption {Random forest regression estimations $\widehat{\sigma(\mathbb{E}_{h}^{d})}$ for $h$ 1NN-DTW, achieving an average improvement in MAE of $37.44\%$ on $D_{val}$ and $37.31\%$ on $D_{test}$ compared to $\ddot{\sigma}_{h}^d$.}
        \label{fig:RF_std}
\end{figure}

We evaluate various regression models on the decomposed multi-target of estimating an algorithm's ($h$) mean performance on a dataset $d$, described by $\mu(\mathbb{E}_{h}^{d})$ and the uncertainty across the k-folds, described by the standard deviation $\sigma(\mathbb{E}_{h}^{d})$. 
The mean performance $\widehat{\mu(\mathbb{E}_{h}^{d})}$ is shown in \Cref{fig:ridge_mean} and the estimated uncertainty $\widehat{\sigma(\mathbb{E}_{h}^{d})}$ is shown in \Cref{fig:RF_std} for $h$ 1NN-DTW (exemplarily). 
Algorithms ridge and random forest are selected based on their performance on $D_{val}$ shown in the upper half and evaluated on $D_{test}$ shown in the lower half. Reporting average relative improvement to account for the different baseline levels.

The hatched area indicates predictions with a relative improvement, while un-hatched areas cover points where the performance estimation is further off than the naive baseline. Our performance estimation achieves a relative improvement, if $|{\widehat{\mu(\mathbb{E}_{h}^{d})} - \mu{(\mathbb{E}_{h}^{d}})}| < |{\ddot{\mu}_{h}^d -\mu{(\mathbb{E}_{h}^{d}})}|$.
This dynamic applies both to comparing the ground truth mean performance to the estimated mean performance $\widehat{\mu(\mathbb{E}_{h}^{d})}$ as well as the ground truth standard deviation across the k-folds compared to the estimated uncertainty $\widehat{\sigma(\mathbb{E}_{h}^{d})}$.
Algorithms ridge and random forest are selected based on their performance on $D_{val}$ shown in the upper half and evaluated on $D_{test}$ shown in the lower half.

\subsection{Estimation improvements benchmark}\label{sec:estimation_regressors}

In \Cref{tab:relative_improvement_main_paper}, we report the absolute error level on the test set as the mean absolute error (MAE) and the relative improvement compared to the naive baseline. For each algorithm (row) we report:
\textbf{(1)} Which model is selected based on its $MAE$ performance on the validation set. \textbf{(2)} The performance of naive baselines $\ddot{\mu}_{h}^d$ and $\ddot{\sigma}_{h}^d$. \textbf{(3)} $MAE$ in predicting the mean performance $\widehat{\mu(\mathbb{E}_{h}^{d})}$ and the std. across the k-folds $\widehat{\sigma(\mathbb{E}_{h}^{d})}$. \textbf{(4)} The relative change of $MAE$ in $\%$.

\begin{table}[tbh]
\centering
\resizebox{0.95\linewidth}{!}{
         \begin{tabular}{l|c|c|c|cHHH||c|c|c|cHHH}
        \toprule
         & \multicolumn{7}{l||}{\textbf{Mean $\mu$}} & \multicolumn{7}{l}{\textbf{Std. $\sigma$}} \\
         &  & \multicolumn{3}{l}{\textbf{MAE}} & & & & & \multicolumn{3}{|l}{\textbf{MAE}} &  \\
          $h(.)$ & $r(.)$ & $\ddot{\mu}_{h}^d$ & $\widehat{\mu(\mathbb{E}_{h}^{d})}$ & $\Delta \%$ & $\ddot{\mu}_{h}^d$ & $\widehat{\mu(\mathbb{E}_{h}^{d})}$ & $\Delta \%$ & $r(.)$ & $\ddot{\sigma}_{h}^d$ & $\widehat{\sigma(\mathbb{E}_{h}^{d})}$ & $\Delta \%$& $\ddot{\sigma}_{h}^d$ & $\widehat{\sigma(\mathbb{E}_{h}^{d})}$ & $\Delta \%$\\
        \midrule
        1NN-DTW & Ri & 0.1277 & 0.1039 & \textbf{-18.61} & 0.0275 & 0.0189 & \textbf{-31.45} & RF & 0.011 & 0.0069 & \textbf{-37.31} & 0.0002 & 0.0001 & \textbf{-48.98} \\
        Arsenal & RF & 0.1046 & 0.1039 & \textbf{-0.72} & 0.0191 & 0.0221 & 15.75 & GB & 0.0094 & 0.0105 & 12.23 & 0.0001 & 0.0002 & 11.97 \\
        BOSS & Ri & 0.1242 & 0.1099 & \textbf{-11.55} & 0.0235 & 0.0213 & \textbf{-9.46} & GB & 0.0134 & 0.0104 & \textbf{-22.68} & 0.0003 & 0.0002 & \textbf{-12.78} \\
        CIF & RF & 0.1072 & 0.1152 & 7.44 & 0.0203 & 0.0237 & 16.61 & GB & 0.0125 & 0.0085 & \textbf{-32.07} & 0.0002 & 0.0001 & \textbf{-50.87} \\
        CNN & Ri & 0.1786 & 0.144 & \textbf{-19.37} & 0.044 & 0.0307 & \textbf{-30.19} & AB & 0.0184 & 0.0149 & \textbf{-19.02} & 0.0005 & 0.0004 & \textbf{-25.71} \\
        Catch22 & Ri & 0.1205 & 0.098 & \textbf{-18.67} & 0.0228 & 0.0164 & \textbf{-28.15} & GB & 0.0122 & 0.0083 & \textbf{-31.82} & 0.0002 & 0.0001 & \textbf{-41.65} \\
        DrCIF & GB & 0.1054 & 0.1 & \textbf{-5.09} & 0.017 & 0.0171 & 0.44 & RF & 0.011 & 0.0086 & \textbf{-21.69} & 0.0002 & 0.0001 & \textbf{-30.30} \\
        EE & Ri & 0.1159 & 0.1013 & \textbf{-12.64} & 0.0221 & 0.0164 & \textbf{-26.08} & RF & 0.0127 & 0.0103 & \textbf{-18.57} & 0.0002 & 0.0002 & \textbf{-32.05} \\
        FreshPRINCE & Ri & 0.1157 & 0.095 & \textbf{-17.92} & 0.0199 & 0.0153 & \textbf{-23.37} & GB & 0.0122 & 0.0089 & \textbf{-27.19} & 0.0002 & 0.0001 & \textbf{-36.80} \\
        HC1 & AB & 0.1055 & 0.1017 & \textbf{-3.60} & 0.019 & 0.022 & 16.10 & RF & 0.0109 & 0.0103 & \textbf{-5.58} & 0.0002 & 0.0002 & 6.42 \\
        HC2 & RF & 0.0957 & 0.0905 & \textbf{-5.45} & 0.0166 & 0.0189 & 14.24 & RF & 0.0104 & 0.0102 & \textbf{-1.82} & 0.0001 & 0.0002 & 14.84 \\
        Hydra-MR & RF & 0.0954 & 0.0925 & \textbf{-2.95} & 0.0172 & 0.0189 & 9.96 & RF & 0.0101 & 0.0106 & 4.70 & 0.0002 & 0.0002 & 13.16 \\
        Hydra & KN & 0.1058 & 0.0966 & \textbf{-8.70} & 0.0186 & 0.0193 & 3.67 & RF & 0.0104 & 0.01 & \textbf{-4.08} & 0.0002 & 0.0002 & 10.39 \\
        InceptionT & RF & 0.0943 & 0.0901 & \textbf{-4.52} & 0.0162 & 0.0182 & 11.78 & RF & 0.0108 & 0.0084 & \textbf{-22.77} & 0.0002 & 0.0002 & \textbf{-22.18} \\
        Mini-R & RF & 0.1021 & 0.1029 & 0.77 & 0.018 & 0.0224 & 24.15 & RF & 0.0098 & 0.0091 & \textbf{-7.51} & 0.0002 & 0.0002 & \textbf{-0.47} \\
        MrSQM & RF & 0.1116 & 0.1112 & \textbf{-0.37} & 0.0205 & 0.0252 & 22.97 & RF & 0.0139 & 0.0132 & \textbf{-4.80} & 0.0003 & 0.0004 & 13.99 \\
        Multi-R & Ri & 0.0973 & 0.0878 & \textbf{-9.76} & 0.0176 & 0.0157 & \textbf{-10.51} & GB & 0.0104 & 0.0104 & 0.12 & 0.0002 & 0.0002 & \textbf{-3.61} \\
        PF & Ri & 0.1134 & 0.0911 & \textbf{-19.64} & 0.0204 & 0.015 & \textbf{-26.73} & RF & 0.0123 & 0.0104 & \textbf{-16.00} & 0.0002 & 0.0002 & \textbf{-12.33} \\
        RDST & RF & 0.1001 & 0.1037 & 3.63 & 0.0188 & 0.0223 & 18.90 & AB & 0.0099 & 0.0105 & 6.65 & 0.0001 & 0.0002 & 30.59 \\
        RISE & Ri & 0.1429 & 0.1241 & \textbf{-13.14} & 0.0307 & 0.0307 & 0.01 & GB & 0.011 & 0.0078 & \textbf{-28.85} & 0.0002 & 0.0001 & \textbf{-43.21} \\
        ROCKET & KN & 0.1019 & 0.094 & \textbf{-7.80} & 0.0182 & 0.0173 & \textbf{-4.96} & RF & 0.0093 & 0.0093 & \textbf{-0.38} & 0.0001 & 0.0001 & \textbf{-3.39} \\
        RSF & KN & 0.1327 & 0.1227 & \textbf{-7.56} & 0.0257 & 0.0228 & \textbf{-11.37} & RF & 0.0127 & 0.0067 & \textbf{-46.95} & 0.0002 & 0.0001 & \textbf{-64.36} \\
        RSTSF & AB & 0.1008 & 0.0943 & \textbf{-6.47} & 0.015 & 0.0157 & 4.90 & GB & 0.0113 & 0.011 & \textbf{-2.71} & 0.0002 & 0.0002 & \textbf{-6.64} \\
        ResNet & AB & 0.1214 & 0.1096 & \textbf{-9.71} & 0.0237 & 0.0247 & 4.19 & KN & 0.0167 & 0.0111 & \textbf{-33.25} & 0.0004 & 0.0002 & \textbf{-50.11} \\
        STC & Ri & 0.1136 & 0.1051 & \textbf{-7.42} & 0.0209 & 0.021 & 0.77 & AB & 0.0132 & 0.0143 & 8.47 & 0.0003 & 0.0006 & 128.75 \\
        STSF & RF & 0.1143 & 0.1132 & \textbf{-0.89} & 0.0189 & 0.0208 & 10.00 & GB & 0.0118 & 0.0088 & \textbf{-25.07} & 0.0002 & 0.0001 & \textbf{-36.60} \\
        ShapeDTW & Ri & 0.1562 & 0.1252 & \textbf{-19.86} & 0.0331 & 0.0215 & \textbf{-35.21} & GB & 0.0106 & 0.0066 & \textbf{-37.79} & 0.0002 & 0.0001 & \textbf{-57.41} \\
        Signatures & GB & 0.1342 & 0.107 & \textbf{-20.28} & 0.0273 & 0.0196 & \textbf{-28.31} & RF & 0.0121 & 0.0084 & \textbf{-30.12} & 0.0002 & 0.0001 & \textbf{-45.03} \\
        TDE & KN & 0.1086 & 0.1095 & 0.88 & 0.0196 & 0.021 & 6.69 & Ri & 0.0139 & 0.0111 & \textbf{-19.73} & 0.0003 & 0.0002 & \textbf{-25.37} \\
        TS-CHIEF & AB & 0.1009 & 0.1113 & 10.20 & 0.0177 & 0.023 & 29.68 & GB & 0.0111 & 0.0125 & 12.84 & 0.0002 & 0.0002 & 40.14 \\
        TSF & Ri & 0.1324 & 0.113 & \textbf{-14.59} & 0.0269 & 0.0214 & \textbf{-20.62} & RF & 0.0137 & 0.0078 & \textbf{-43.37} & 0.0002 & 0.0001 & \textbf{-61.57} \\
        TSFresh & KN & 0.1501 & 0.1349 & \textbf{-10.12} & 0.0315 & 0.0257 & \textbf{-18.57} & GB & 0.0405 & 0.0338 & \textbf{-16.49} & 0.0034 & 0.003 & \textbf{-11.52} \\
        WEASEL-D & RF & 0.109 & 0.1052 & \textbf{-3.44} & 0.0211 & 0.0247 & 17.16 & RF & 0.0091 & 0.009 & \textbf{-1.73} & 0.0001 & 0.0001 & 4.76 \\
        WEASEL & RF & 0.1199 & 0.1141 & \textbf{-4.86} & 0.023 & 0.0242 & 5.09 & RF & 0.0126 & 0.0101 & \textbf{-19.91} & 0.0002 & 0.0002 & \textbf{-15.33} \\
        cBOSS & AB & 0.1248 & 0.1328 & 6.42 & 0.0244 & 0.031 & 27.22 & Ri & 0.0134 & 0.0108 & \textbf{-18.99} & 0.0003 & 0.0002 & \textbf{-33.89} \\
        \hline
        \textbf{Mean} & - & 0.1167 & 0.1073 & \textbf{-7.32} & 0.0222 & 0.0213 & \textbf{-1.28} & - & 0.0127 & 0.0106 & \textbf{-15.81} & 0.0003 & 0.0003 & \textbf{-14.20} \\
        \bottomrule
        \end{tabular}
}
\caption{Applying AdaBoost (AB), GradientBoosting (GB), KNeighbors (KN), RandomForest (RF), Ridge (RID) as regressors $r(.)$ in our multi-target regression AS problem for all $35$ algorithms $h(.)$ referenced in the benchmark~\parencite{middlehurst2023bake}. Error measured by $MAE$ and relative performance improvements (lower better $\Downarrow$) for both mean performance and uncertainty estimation. Improvements highlighted \textbf{bold}.}
\label{tab:relative_improvement_main_paper}
\end{table} 

\begin{figure}[tbh]
\includegraphics[width=\linewidth]
{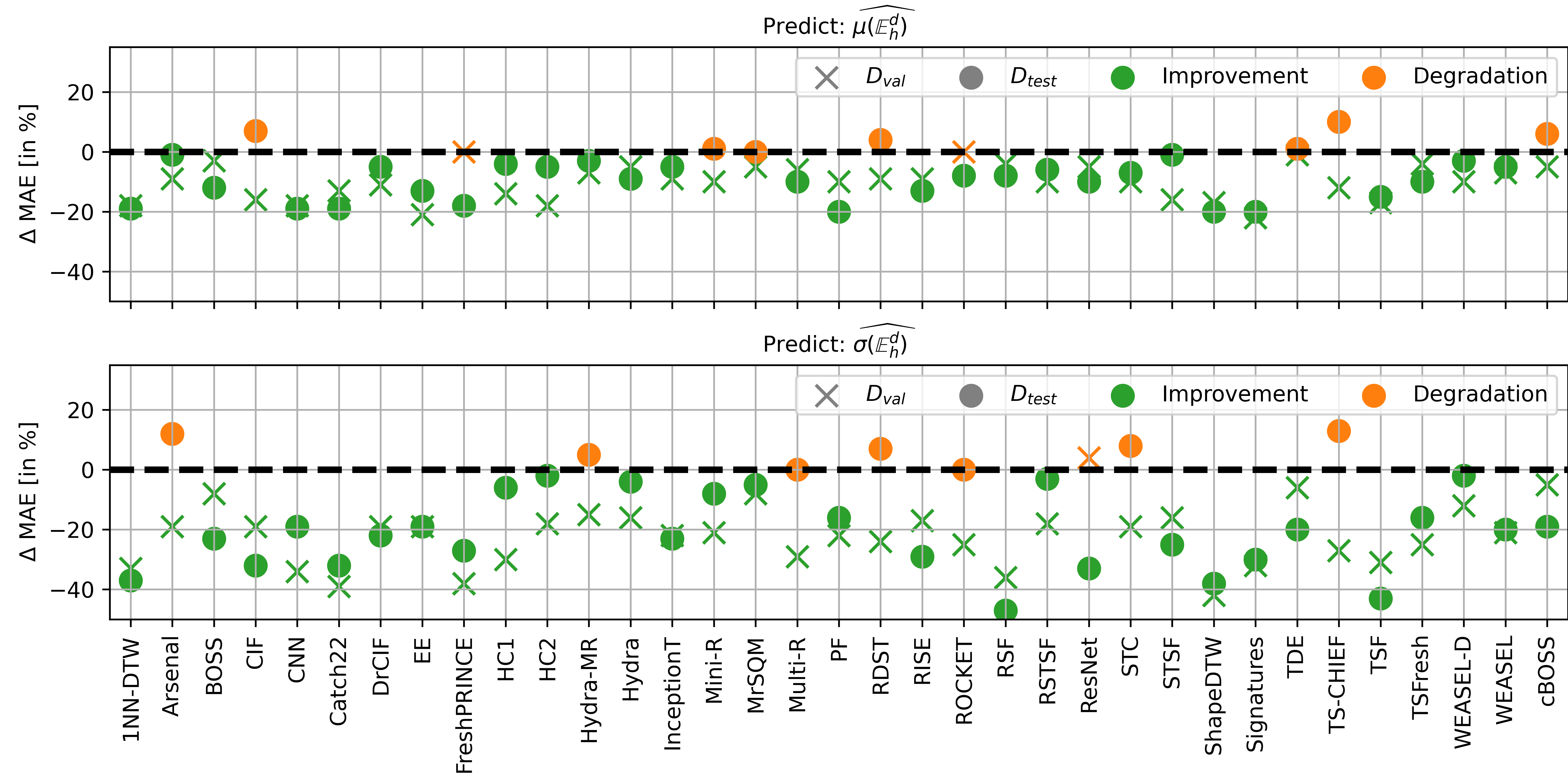}
  \caption{Performance improvement for all $35$ algorithm $h(.)$ in predicting the mean performance $\widehat{\mu(\mathbb{E}_{h}^{d})}$ and uncertainty $\widehat{\sigma(\mathbb{E}_{h}^{d})}$ on validation and test set. Reporting the relative change compared to the naive baseline $\ddot{\mu}_{h}^d$ and $\ddot{\sigma}_{h}^d$ (lower better). Algorithms on the x-axis are sorted by name.}
  \label{fig:test_data_improvement}
\end{figure}

For example, when a ridge regressor estimates the performance of the \textit{1NN-DTW} algorithm, as shown in \Cref{tab:relative_improvement_main_paper}, we observe a significant improvement: Just by analyzing the fingerprint $f_D^d(.)$ of the unseen test datasets, our approach is $18.61\%$ more accurate in estimating the mean ground truth performance on these datasets $\mathbb{E}_{h}^{d}$, compared to the naive baseline measured by the MAE.
Estimating the uncertainty of the \textit{1NN-DTW}-algorithm $\widehat{\sigma(\mathbb{E}_{h}^{d})}$ our approach improves the naive baseline by $37.31 \%$ in MAE.
A visual interpretation of these results as well as a comparison of the performance on the validation datasets $D_{val}$ and the test datasets $D_{test}$ is given in \cref{fig:test_data_improvement}. 
In summary, based on the validation set performance for each algorithm $h(.)$, our approach outperforms the naive baseline by an average of $-7.32 \%$ for the MAE when predicting the mean performance and by $-15.81 \%$ in the MAE of the std. deviation.
Our approach is capable of differentiating performances on individual datasets. 
Instead of selecting an algorithm based on its average performance on some publicly available benchmark, our approach allows for precise estimation of the exact performance each algorithm will achieve on a specific dataset. 
This enables a more informed and tailored algorithm selection process, ensuring that the chosen algorithm is the most suitable for the unique characteristics of the new dataset.

\section{Limitations \& Future Work}

While our current approach demonstrates significant advancements, it also has certain limitations that opens various directions for future work.
The decomposition of the multi-target regression overlooks the intricate dependencies between algorithms and the collective objectives \parencite{lorena2008review}. In future work, this can be investigated by a regressor that predicts the performance of multiple algorithms at once. 
The selection requires domain experts to balance different objectives, such as mean performance and uncertainty, which can complicate the AS process. 
Further development into an AI service could incorporate relaxations of multiple objectives and include domain experts in a human-in-the-loop manner. 
Our approach tests a range of regressors, including those with transparent internal mechanics, but does not inherently prioritize regressors based on their interpretability. This may restrict its usefulness when understanding a model's internal decision-making process, which is often crucial in real-world applications. The predictions and properties, such as the interpretability, could be combined in a decision rule for the final AS that meets the user's preferences. 
The effectiveness of our estimation strategy depends on the chosen metric, with MAE showing different levels of robustness and volatility in performance improvements across validation and test sets (as documented in \Cref{tab:relative_improvement_main_paper}). To improve the robustness of our approach, we encourage researchers to share their data fingerprint and the corresponding performances \parencite{koester2020panel}. 
For the next steps, our performance estimations can guide service providers in the domain, so instead of auto-correcting human decisions, we can provide feedback on which algorithm would be more suited \parencite{balla2023feeding}. Such an extension follows the trajectory of leveraging technology to advance service, as suggested by \cite{ostrom2010moving}.
Further developments of our approach can follow up on the ongoing discussion about which additional objectives to assess (e.g., expected running time \parencite{bossek2019multi}). Our adaptable and extensible approach allows us to estimate such objectives instead or aside from the performance and uncertainty. 

\section{Conclusion}
This paper introduces a novel data fingerprint for time-series classification, offering an approach to support more effective and privacy-preserving AI development without having access to all data points. We predict algorithmic performance and associated uncertainties by strategically decomposing the multi-target regression problem.

Our assessment across $112$ datasets of the University of California riverside benchmark showcases its capability in accurately forecasting the outcomes of $35$ state-of-the-art algorithms, surpassing a naive baseline by an average of $7.32\%$ in estimating mean performance and $15.81\%$ in quantifying uncertainty. 
Our approach will assist researchers and professionals in the field of algorithm selection in time series classification to set up successful AI services. %
We encourage other researchers and practitioners to use and extend the approach with the proposed fingerprints for further objectives. A promising field of research lies ahead.

\printbibliography

\end{document}